\documentclass[sigconf,nonacm,natbib=true]{acmart}

\usepackage{graphicx}
\usepackage{amsmath}
\usepackage{booktabs}
\usepackage{multirow} 
\usepackage{listings}
\usepackage{tabularx}

\usepackage{soul} % for highlighting

\AtBeginDocument{%
  }

\renewcommand\footnotetextcopyrightpermission[1]{}
\begin{document}

\title{Chain of Summaries: Condensing Information by~Anticipating ~Questions}

\author{William Brach}
\affiliation{%
  \institution{Slovak Technical University}
  \country{Slovakia}}
\email{william.brach@stuba.sk}

\author{Kristian Koštáľ}
\affiliation{%
  \institution{Slovak Technical University}
  \country{Slovakia}}
\email{kristian.kostal@stuba.sk}

\author{Lukas Galke Poech}
\affiliation{%
  \institution{University of Southern Denmark}
  \country{Denmark}}
\email{galke@imada.sdu.dk}

\renewcommand{\shortauthors}{Brach et al.}


\begin{abstract}
Large Language Models (LLMs) are increasingly using external web content.
However, much of this content is not easily digestible by LLMs due to LLM-unfriendly formats and limitations of context length. To address this issue, we propose a method for generating general-purpose, information-dense summaries that act as plain-text repositories of web content. Inspired by Hegel's dialectical method, our approach, denoted as Chain of Summaries (CoS), iteratively refines an initial summary (thesis) by identifying its limitations through questioning (antithesis), leading to a general-purpose summary (synthesis) that can satisfy current and anticipate future information needs. Experiments on the TriviaQA, TruthfulQA, and SQUAD datasets demonstrate that CoS outperforms zero-shot LLM baselines by up to 66\% and specialized summarization methods such as Chain of Density, BRIO and PEGASUS by up to 27\%. CoS-generated summaries yield higher Q\&A performance compared to the source content, while requiring substantially fewer tokens and being agnostic to the specific downstream LLM. CoS thus resembles an appealing option for website maintainers to make their content more accessible for LLMs, while retaining possibilities for human oversight.
\end{abstract}

 

\maketitle

\section{Introduction}
\begin{figure}[ht!]
    \centering
    \includegraphics[width=\columnwidth]{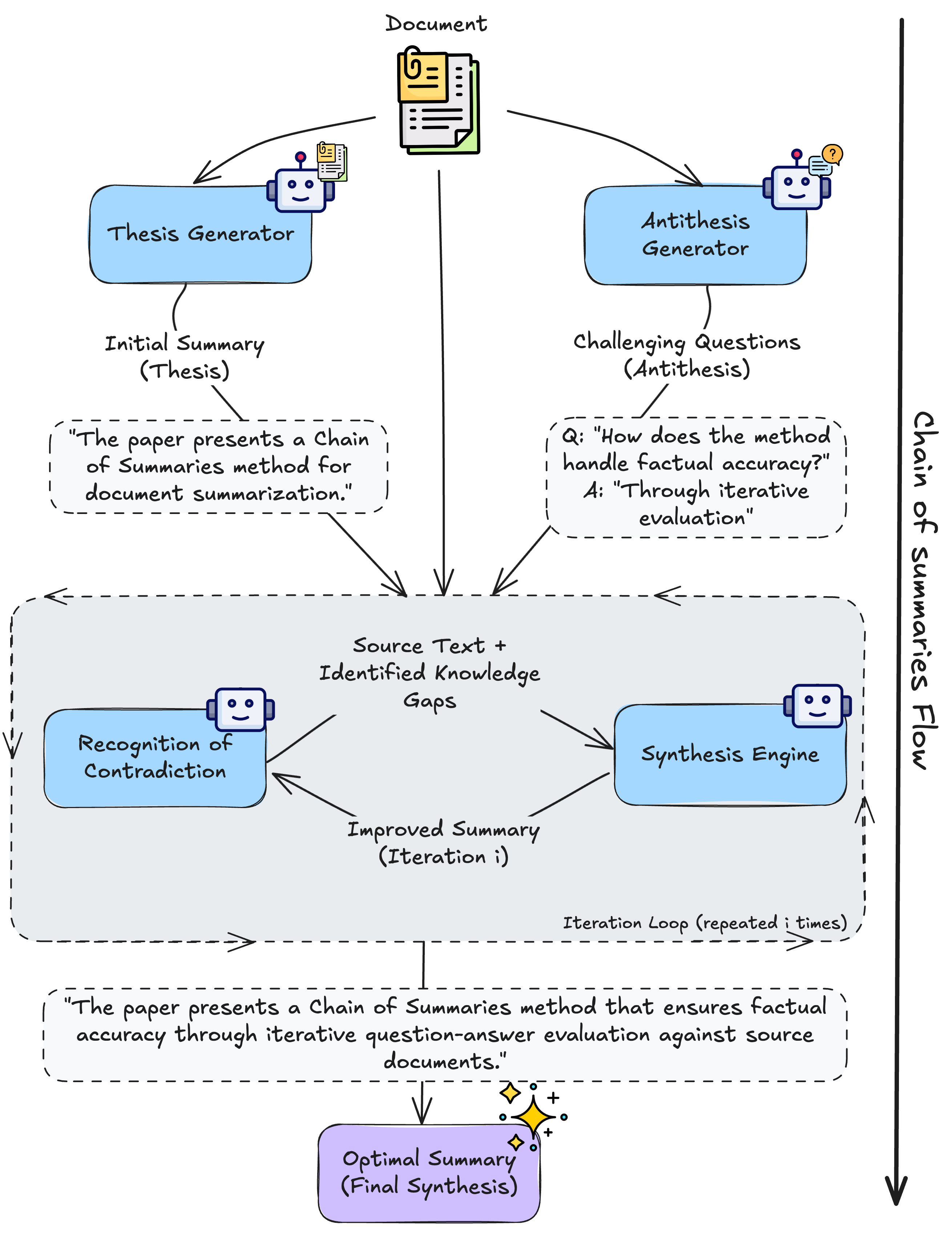}
    \caption{Chain of Summaries (CoS) method implements Hegel's dialectical method for document summarization, iteratively refining an initial summary (thesis) through challenging questions (antithesis) to produce an information-dense optimal summary (synthesis) capable of answering diverse queries.}\label{fig:chain-of-summaries}
\end{figure}

Large language models (LLMs) are given increasing autonomy to navigate and process information on the Web (e.g., Search and Deep Research functionalities in commercial systems). However, most of the web content is not easy to digest for LLMs and challenges their context size, leading to necessary compromises in performance and potential instabilities. It has been proposed to make the content on the web more accessible to LLMs by running parallel versions of the website in an LLM-friendly markdown format~\cite{howard2024llmstxt}. But how can we turn the plethora of existing content on the Web into information-dense chunks of plain text that capture the essence of the original content along various, potentially unpredictable, aspects? 

Here, we look at this problem from the angle of summarization in the sense of making information better available for LLMs to consume. Concretely, if we had a robust summarization method that captured the essence of the content in various aspects, we could hope that a variety of future questions could be answered just based on the summary. If questions come up that cannot be answered from the summary, the full website could be consulted, and the summary could be revised accordingly for future requests. This way, the plain-text summary would effectively act as a cache for websites, which can be implemented both server-side (e.g., website maintainers) and client-side (e.g., Agentic systems) and can be subject to human oversight as needed.

The key idea of this paper is to frame this process through the lens of Hegel's dialectical method~\cite{sep-hegel-dialectics} and implement the procedure with multiple LLM calls. Our main hypothesis is that an initial summary draft (the thesis in Hegel's terms) can be iteratively refined based on its capability to answer synthetically generated questions (the antithesis), creating a general-purpose summary (synthesis) that facilitates answering as many unseen future questions as possible -- thereby, maximizing the number of cache hits.
We denote this procedure as the Chain of Summaries (CoS; Figure~\ref{fig:chain-of-summaries}).

Representing the summary in natural language has two key advantages: First, it ensures human verifiability; website maintainers must be able to easily read and validate the information that LLM agents will consume, preventing the spread of inaccurate or outdated content. Second, a plain-text format is model-agnostic, creating a universal and durable representation that any LLM agent can use, in contrast to model-specific formats like dense embeddings from language models that are often used for retrieval-augmented generation~\cite{lewis2020retrieval}.

To test the effectiveness of our Chain of Summaries approach, we focus our experiments on question-answering datasets. This is because considering a question-answering task enables us to take questions from the dataset that we can run against the summary of the content for evaluation. This further allows us to compare how helpful synthetically generated questions are compared to questions taken from the training split of these datasets (human-crafted). 

With the aim of a representative selection, our experiments cover a commercial model (GPT-4o-mini \cite{openai2024gpt4omini}) as well as publicly available models of smaller sizes: Llama-3.2:3B \cite{grattafiori2024llama3herdmodels} and Qwen-2.5:7B \cite{qwen2.5} to enact the different roles in the CoS framework: (i) generate the initial summary; (ii) synthesize questions;  (iii) determine whether they can be answered from the summary; and (iv) iteratively refine the summary. See Figure~\ref{fig:chain-of-summaries} for an overview of the procedure.
Although technically, models can be mixed and matched, we keep the LLM backbone constant for our experiments. It is expected that GPT-4o-mini will perform best, yet we are interested in how much performance can be retained using smaller, local models.

We further experimented with several different hyperparameters of the CoS approach and found that iteratively refining the summary (e.g., iterate 10 times with 1 question per iteration) has a clear advantage over presenting all questions at once (e.g., 50 questions). Notably, we split available questions into training, validation, and test sets, and selected the best-performing summary for each document based on validation performance before evaluating it on the test set.

A key factor of interest is to what extent the training set of questions, that is available for question-answering datasets but not for summarizing arbitrary content on the web, can be replaced by having another language model generating synthetic questions. We will show that synthetic questions are on par or almost as good in all cases.

In general, our experiments show that our method of creating information-dense summaries through generating synthetic questions and iteratively refining the summary is highly effective: It shortens the content but also makes the core information better accessible to LLMs. We find that it often leads to an even higher question-answering performance than when using the full source document as input. Chain of Summaries also outperforms competing summarization techniques, traditional ones such as BRIO and PEGASUS, or modern ones based on zero-shot prompting a language model for a summary, or other iterative approaches such as Chain of Density.

Finally, we conduct a break-even analysis to study after how many queries to an assumed website the upfront cost of the summarization procedure is amortized. We find that it takes only 13-15 queries for which the summary can be consulted instead of the full document.

In sum, this paper makes three core contributions:

% Finally, we investigate how much the multi-step CoS summarization method can be reduced to a single step. For this, we first fine-tune an encoder-only model to estimate the quality of downstream QA performance (through regression on the main evaluation metric). Then we use this encoder-only model as a reward model for reinforcement fine-tuning Qwen2:0.5B-Instruct~\cite{qwen2} and Qwen3:0.6B~\cite{qwen3} via Group Relative Policy Optimization (GRPO)~\cite{shao2024deepseekmathpushinglimitsmathematical}. Our results show that CoS consistently outperforms the strong zero-shot summarization baseline, given the same backbone LLM -- suggesting that CoS improves summaries' QA performance. Our reinforcement-finetuned Qwen2:0.5B-Instruct and Qwen3:0.6B models performed slightly worse than the pure CoS approach, indicating that the incremental procedure is beneficial to obtain a high-quality summary with the limitation of potential reward hacking. Similarly, CoS outperforms the established specialized summarization baselines, such as BRIO~\cite{liu2022briobringingorderabstractive} and PEGASUS~\cite{zhang2019PEGASUS}, as well as strong zero-shot summaries generated by LLM.

\begin{itemize}

    \item We propose and evaluate the Chain of Summaries (CoS) approach, which is based on Hegel's dialectical method. In this CoS, summaries serve as machine-consumable-human-readable knowledge caches, and quality is measured by downstream QA performance.
    \item We provide empirical evidence that the proposed method improves upon the naive-but-strong baseline of zero-shot summarization with an LLM, and outperforms strong dedicated summarization baselines.
    \item We provide a efficiency analysis showing that CoS summaries reduce per-query token consumption by over 98\%, with break-even occurring after approximately 14 queries, making CoS cost-effective for repeatedly-accessed documents.
\end{itemize}

\section{Related Work}

Text summarization has seen advances from both abstractive and extractive single-pass methods. Abstractive models include PEGASUS~\cite{10.1016/j.neucom.2024.128280,zhang2023benchmarkinglargelanguagemodels}, with its sentence-reconstruction pre-training, and BRIO~\cite{Gambhir2017}, which improves candidate summary ranking. In extractive summarization, recent work like DiffuSum~\cite{zhang-etal-2023-diffusum} uses diffusion models to match summary representations with source sentences. Summarization via large language models is becoming increasingly popular, but often suffers from the "middle curse"~\cite{ravaut-etal-2024-context,liu2023lostmiddlelanguagemodels}, a strong positional bias where models struggle to utilize information from the middle of long contexts.

Recent work has focused on improving LLM-retrieval integration through two main approaches. The first treats LLMs as black boxes while refining retrieved context: Replug \cite{shi-etal-2024-replug} uses perplexity scores to train the retriever, Fit-Rag \cite{10.1145/3676957} employs bi-label scoring for relevance and LLM preferences, and Recomp \cite{xu2023recompimprovingretrievalaugmentedlms} compresses documents to reduce noise.

The second approach fine-tunes LLMs for better knowledge integration. Atlas \cite{10.5555/3648699.3648950} jointly pre-trains retriever and generator, while Raven \cite{huang2024ravenincontextlearningretrievalaugmented} introduces Fusion-in-Context Learning. Ra-Dit \cite{lin2024raditretrievalaugmenteddualinstruction} proposes dual instruction tuning, ChatQA \cite{liu2024chatqasurpassinggpt4conversational} uses two-stage instruction tuning, and Rankrag \cite{10.5555/3737916.3741766} unifies ranking and generation in a single model. Genki \cite{shen2025genkienhancingopendomainquestion} integrates knowledge directly into model parameters via fine-tuning.

To address limitations of single-pass generation, iterative refinement approaches have emerged. SELF-REFINE~\cite{10.5555/3666122.3668141} enables LLMs to generate, critique, and revise their outputs, while SummIt~\cite{zhang2023summit} combines summarization with self-evaluation to improve faithfulness without supervised training. SuRe~\cite{kim2024sure} generates multiple answer summaries and uses the summarization process itself as a verification mechanism.

Chain of Density~\cite{adams2023sparsedensegpt4summarization} iteratively densifies summaries while maintaining fixed length, optimizing for entity density with human preference evaluation; in contrast, our CoS approach optimizes for downstream QA performance through dialectical refinement. Building on these advances, our work introduces a question-guided iterative framework that explicitly optimizes for information density by ensuring summaries can answer the most important questions about the source text, bridging QA-based evaluation with retrieval-augmented generation and iterative refinement.

\section{Chain of Summaries}\label{sec:cos}

The Chain of Summaries (CoS) approach is inspired by Hegel's dialectical method, whereby knowledge evolves through the stages of \textbf{thesis}, \textbf{antithesis}, and \textbf{synthesis}. As shown in Figure~\ref{fig:chain-of-summaries}, our system works in phases with several interconnected components that progressively enhance the summaries' quality. 

Given a source document $D$, our method generates a set of question-answer pairs $Q = \{(q_1, a_1), (q_2, a_2), \ldots, (q_n, a_n)\}$ and partitions them into training, validation, and test sets: $Q = Q_{\text{train}} \cup Q_{\text{val}} \cup Q_{\text{test}}$. The iterative refinement process operates as follows:

\textbf{Initial Summary Generation (Thesis):} We first use an LLM to generate a zero-shot summary $s^0$ from the source document $D$. This initial thesis provides a preliminary   understanding that inevitably contains internal contradictions and limitations.

\textbf{Question Generation (Antithesis):} Using the prompt in Listing~\ref{lst:qa_generation}, we generate the synthetic question-answer set $Q$ from $D$. These questions act as the antithesis by challenging the summary's claim to comprehensiveness, highlighting areas where key information is omitted.

\textbf{Iterative Refinement Process (Movement towards Synthesis):} For each iteration $k \in \{1, \ldots, i\}$, the dialectical cycle proceeds as:

\begin{itemize}
  \item \textbf{Evaluation Phase:} We evaluate the current summary $s^{k-1}$ against questions in $Q_{\text{train}}$ using the prompt in Listing~\ref{lst:answer_prompt}. Let
  $U^{k} \subseteq Q_{\text{train}}$ denote the set of \emph{unanswered questions}—those for which $s^{k-1}$ fails to provide correct answers. These reveal the thesis's limitations.
  \item \textbf{Refinement Phase:} We sample a subset of $iq$ questions from $U^{k}$ and invoke a synthesis engine that takes as input the tuple $(D, s^{k-1}, U^{k}_{iq})$ and produces an improved summary $s^{k}$ that incorporates the missing information while maintaining conciseness. The prompt for this phase is displayed in Listing~\ref{lst:refine_summary_prompt}.
\end{itemize}

\textbf{Continued Dialectical Progression (Iteration):} This evaluate-refine cycle repeats for $i$ iterations, producing a sequence of candidate summaries $\mathcal{S} = \{s^0, s^1, \ldots, s^i\}$. Each iteration represents a dialectical movement towards the best summary that maximizes information density for prospective downstream question-answering tasks. The final summary $s^*$ is selected based on validation performance.

\paragraph{Summary selection based on validation performance}

Since the iterative refinement process produces multiple candidate summaries (one per iteration), we employ a validation-based selection strategy to choose the final summary for each document. Specifically, we partition the available questions into training, validation, and test sets. The training questions guide the refinement process, while validation questions, never seen during refinement, are used to select the best performing summary across all iterations. Formally, for each document with candidate summaries $\mathcal{S} = \{s^0, s^1, \ldots, s^i\}$ from $i$ iterations, we select:
  
\begin{equation}
  s^* = \underset{s^k, k \in \{1, \ldots, i\}}{\operatorname{arg\,max}} \text{MeanF1}(s^k, Q_{\text{val}})
\end{equation}

where $Q_{\text{val}}$ denotes the validation question set. The selected summary $s^*$ is then evaluated on the held-out test set $Q_{\text{test}}$ to compute the final generalization score. This procedure ensures that our reported results reflect true out of sample performance, preventing overfitting to the questions used during iterative refinement.

\paragraph{Hyperparameters}

The synthesis process is controlled by two key hyperparameters: the number of iterations $i$, which determines how many refinement cycles occur, and the number of refinement questions $iq$, which controls how many unanswered questions from $U^k$ guide each iteration. Higher $i$ enables more comprehensive evolution but risks over-refinement, while $iq$ balances focused improvement (low values) against broader coverage (high values).

\section{Experimental Setup}\label{sec:setup}

\subsection{Datasets}

To directly assess our primary objective of creating summaries maximally useful for downstream QA tasks, we selected standard QA datasets. By testing question-answering using only the summary, we measure information density and utility. Our evaluation uses TriviaQA \cite{2017arXivtriviaqa}, TruthfulQA \cite{lin2022truthfulqameasuringmodelsmimic}, and SQuAD v1.1 \cite{rajpurkar-etal-2016-squad}. We repurpose these datasets by extracting (document, question, answer) triples. This allows us to test if the summary retains enough information to answer the questions, replacing standard reading comprehension tests. To ensure a valid evaluation of our iterative approach, we select documents containing at least three test questions, providing sufficient representation to assess performance across iterations. All compared methods receive identical (document, question, answer) triples, ensuring fair evaluation.

\subsection{LLM Backbones}
For experimental consistency and to isolate the effects of our iterative refinement process, we used the same LLM across all components (initial summary generation, QA pair generation, evaluation, and refinement) within each experimental run. This eliminated confounding variables from mixing models with different capabilities. We evaluated three language models: \textit{GPT-4o-mini}, \textit{Llama3.2:3B}, and \textit{Qwen2.5:7B} for CoS.

\subsection{Baselines}

To provide context for the performance of our CoS approach, we compare it with a range of benchmarks designed to evaluate its effectiveness against direct LLM applications, alternative dense representations, and other summarization techniques. We use identical LLM backbones to generate a direct, one-pass summary. This is the simplest application of LLMs — \textbf{zero-shot summarization} — and shows how our iterative CoS process improves performance. The \textbf{Full Source Content} baseline represents a source document as context for the downstream QA task. This serves as a benchmark to determine whether our summarization process merely condenses information or actively makes it more accessible and useful for an LLM by mitigating issues such as the 'lost-in-the-middle' effect, for example. \textbf{BRIO} and \textbf{PEGASUS} represent supervised abstractive summarization approaches, enabling comparison between our iterative method and specialised summarization techniques. \textbf{Chain-of-Density} \cite{adams2023sparsedensegpt4summarization} is an iterative prompting technique that progressively increases summary information density through multiple LLM passes.

\subsection{Evaluation Metrics}\label{sec:evaluation}

Given our goal of creating summaries optimised for downstream QA, our evaluation intentionally focuses on functional utility rather than traditional summarization metrics like fluency or coherence. Thus, our primary evaluation metric (Correct-F1) derives from the TriviaQA correctness measure using an F1 score for question-answering performance. The F1 score is computed as $\text{F1} = \frac{2 \cdot \text{precision} \cdot \text{recall}}{\text{precision} + \text{recall}}$, where $\text{precision} = \frac{|P \cap G|}{|P|}$ and $\text{recall} = \frac{|P \cap G|}{|G|}$, with $P$ representing prediction tokens and $G$ representing ground truth tokens. The F1 calculation involves preprocessing prediction and ground truth strings by: converting to lowercase, removing punctuation and articles, replacing underscores with spaces, and standardizing whitespace. These normalized strings are then tokenized by spaces for F1 calculation, with common tokens determining precision and recall. For dataset-level evaluation, let $S = \{s_1, s_2, ..., s_n\}$ be all summaries and $I = \{1, 2, ..., m\}$ be all iterations. For each summary $s_j$ and iteration $i$, we calculate
$\text{MeanF1}(s_j, i) = \frac{1}{|E|}\sum_{e \in E} \text{F1}(s_j^i, e)$

where $E$ is all question-answer pairs and $s_j^i$ is summary $j$ at iteration $i$. We report the best-performing iteration for each summary $\text{Correct-F1}(s_j) = \max_{i \in I}(\text{MeanF1}(s_j, I))$

This procedure enables each summary to achieve its best possible performance, as the optimal iteration may differ between summaries. For instance, summary $s_1$ might achieve its best score at iteration $i=10$, while summary $s_2$ peaks at iteration $i=9$.

For comparability with the literature, we also report standard evaluation measures for the datasets: F1 for TriviaQA and SQuADv1.1. For TruthfulQA, we report ROUGE-L to measure the textual similarity between the generated and ground-truth answers, acknowledging this is a proxy that does not fully capture truthfulness but serves as a consistent evaluation metric in our framework.

\paragraph{Memorization} evaluates the summary's performance (using Correct-F1) on the exact set of questions that were used to guide the iterative refinement process. The goal is to quantify how effectively each refinement step incorporates new information to cover identified knowledge gaps. A high score demonstrates that the iterative mechanism is successfully improving the summary based on the provided feedback, not just that the model is memorizing answers.

\paragraph{Generalization} measure of the final summary's performance on the downstream QA task. We apply Correct-F1 to a completely separate, held-out set of validation questions that were never seen during the summary refinement loops. This evaluates the summary's comprehensive knowledge capture, and a high score indicates that the summary has become a robust and broadly applicable knowledge source, useful for novel queries beyond those used for its creation. To ensure an unbiased evaluation, the set of questions used for the final generalization score was a held-out test set, kept separate from any questions used during the iterative refinement and validation selection loops.

\section{Results}\label{sec:results}

\begin{table*}[htbp]
\centering
\resizebox{\textwidth}{!}{
\begin{tabular}{lcccccc}
\toprule
\textbf{Model} & \multicolumn{2}{c}{\textbf{TriviaQA (F1)}} & \multicolumn{2}{c}{\textbf{TruthfulQA (ROUGE-L)}} & \multicolumn{2}{c}{\textbf{SQUAD1.1 (F1)}} \\
& \textbf{Memorization} & \textbf{Generalization} & \textbf{Memorization} & \textbf{Generalization} & \textbf{Memorization} & \textbf{Generalization} \\
\midrule
\multicolumn{7}{c}{\emph{GPT-4o-mini}}\\
 \emph{Original Content} & - & \textit{0.76} & - & 10 & - & \textbf{0.28} \\
 Chain of summaries w/ GPT-4o-mini (ours) & 0.48 & \textbf{0.80} & 30.34 & \textbf{12.69} & 0.46 & \textit{0.27} \\
 Zero shot summary w/ GPT-4o-mini & 0.43 & 0.73 & 24.04 & \textit{10.67} & 0.41 & \textit{0.27} \\
 Chain of Density & - & 0.73 & - & 9.07 & - & 0.26 \\
 BRIO & - & 0.63 & - & 9.25 & - & 0.16 \\
 PEGASUS & - & 0.72 & - & 8.28 & - & 0.26 \\
 % CoS-Finetuned-Qwen3:0.6B & - & 0.73 & - & 9.55 & - & 0.25 \\
 % CoS-Finetuned-Qwen2:0.5B-Instruct & - & 0.74 & - & 8.77 & - & 0.24 \\
\midrule
\multicolumn{7}{c}{\emph{Llama3.2:3B}}\\
 \emph{Original Content} & - & 0.51 & - & 4.52 & - & \textit{0.19} \\
 Chain of summaries w/ Llama3.2:3B (ours) & 0.39 & \textbf{0.62} & 29.31 & \textbf{14.77} & 0.34 & \textbf{0.20} \\
 Zero shot summary w/ Llama3.2:3B  & 0.21 & \textit{0.52} & 18.34 & \textit{8.89} & 0.23 & 0.18 \\
 Chain of Density & - & 0.46 & - & 4.63 & - & 0.17 \\
 BRIO & - & 0.45 & - & 5.74 & - & 0.14 \\
 PEGASUS & - & 0.50 & - & 3.29 & - & 0.18 \\
  % CoS-Finetuned-Qwen3:0.6B & - & 0.50 & - & 5.33 & - & \textit{0.19} \\
  %   CoS-Finetuned-Qwen2:0.5B-Instruct & - & 0.49 & - & 6.48 & - & 0.18 \\
\midrule
\multicolumn{7}{c}{\emph{Qwen2.5:7B}}\\
 \emph{Original Content} & - & 0.41 & - & 6.25 & - & \textbf{0.16} \\
 Chain of summaries w/ Qwen2.5:7B (ours) & 0.52 & \textbf{0.60} & 46.06 & \textbf{10.71} & 0.59 & \textbf{0.16} \\
 Zero shot summary w/ Qwen2.5:7B  & 0.23 & \textit{0.48} & 18.55 & \textit{7.35} & 0.36 & 0.14 \\
 Chain of Density & - & 0.36 & - & 5.13 & - & 0.12 \\
 BRIO & - & 0.21 & - & 5.74 & - & 0.07 \\
 PEGASUS & - & 0.43 & - & 5.92 & - & 0.12 \\
 % CoS-Finetuned-Qwen3:0.6B & - & 0.45 & - & 4.89 & - & \textit{0.15} \\
 %  CoS-Finetuned-Qwen2:0.5B-Instruct & - & 0.42 & - & 5.33 & - & 0.13 \\
\bottomrule
\end{tabular}%
}
\caption{A performance comparison of summarization methods across three language model backbones on the TriviaQA, TruthfulQA, and SQUAD 1.1 datasets. Our CoS approach outperforms all baselines on generalization metrics across all model backbones and datasets. Bold values indicate the best performance, and italicized values indicate the second-best performance. The table also shows that CoS effectively retains critical information through iterative refinement.}\label{tab:results}
\end{table*}

\begin{table*}[ht]
\small

\centering
\begin{tabularx}{\textwidth}{p{0.45\textwidth} | p{0.50\textwidth}}
\toprule
\textbf{Zero-Shot Summary} & \textbf{Chain of Summaries (CoS) Summary} \\ \midrule
Sherlock Holmes, created by Sir Arthur Conan Doyle, is a renowned fictional detective known for his exceptional observational and deductive skills, first appearing in 1887. & 
Sherlock Holmes, created by Sir Arthur Conan Doyle, is a renowned fictional detective known for exceptional observation and deductive skills, first appearing in 1887. \\
\addlinespace
His character has significantly influenced mystery literature and popular culture, leading to numerous adaptations and a lasting legacy as one of the most portrayed characters in film history. & 
\hl{Holmes requires keen observation skills for deduction.} Conan Doyle cited \hl{Joseph Bell as an inspiration.} \dots \hl{Holmes feels disinterested in personal relationships.} A prevalent theme is \hl{Holmes's distrust of women.} He sometimes uses charm to influence others. The original setting for most investigations is \hl{Victorian London.} \\ 
\addlinespace
& \hl{Robert Downey Jr.} starred in adaptations directed by \hl{Guy Ritchie.} Holmes's \hl{estimated birth year is 1854} based on "His Last Bow." The first series is \hl{The Adventures of Sherlock Holmes.} He fails to recognize unimportant facts. Holmes achieved a \hl{university degree} before detection. "The Hound of the Baskervilles" features \hl{Holmes's return} after death. He employs \hl{abductive reasoning} to solve cases. \\ \bottomrule
\end{tabularx}
\caption{Comparison of Zero-Shot vs. CoS Summarization Density. The Zero-Shot summary captures the \textbf{global context} but misses 85\% of the specific factual entities. The CoS summary maintains the global context while surfacing granular details (biographical dates, inspirations, and specific adaptations) that are critical for high-fidelity information retrieval.}
\label{tab:Comparison_zs_vs_cos}
\end{table*}

We evaluate our CoS approach on the TriviaQA, TruthfulQA, and SQUAD v1.1 datasets. We benchmark CoS against a suite of baselines, including direct LLM outputs, alternative dense representations like structured data, and both classic and state-of-the-art summarization methods. Our evaluation focuses on two key areas: the effectiveness of the refinement process on guiding questions, and more importantly, the generalization performance on a completely separate, held-out set of unseen questions.

Table~\ref{tab:results} presents the main results, and Table~\ref{tab:Comparison_zs_vs_cos} highlights a representative CoS summary versus a zero-shot summary.
CoS consistently outperforms zero-shot summarization across all models and datasets, with improvements using GPT-4o-mini as backbone on TriviaQA (0.80 vs. 0.73, a 9.6\% gain) and in Llama3.2:3B on TruthfulQA (14.77 vs. 8.89, a 66\% improvement). Our approach also exceeds traditional summarization methods, surpassing BRIO by 27\% (0.80 vs. 0.63) and PEGASUS by 11\% (0.80 vs. 0.72) on TriviaQA with GPT-4o-mini. Notably, CoS also outperforms Chain of Density, another iterative prompting approach, by 9.6\% with GPT-4o-mini (0.80 vs. 0.73). This advantage becomes more pronounced with smaller models, where CoS exceeds Chain of Density by 35\% with Llama3.2:3B (0.62 vs. 0.46) and by 67\% with Qwen2.5:7B (0.60 vs. 0.36), suggesting that the question-guided dialectical refinement of CoS is more effective than density-based iteration, particularly when model capacity is limited. A key finding is that CoS summaries consistently present information more effectively for downstream QA purposes than the original documents themselves. For instance, on TriviaQA with a GPT-4o-mini backbone, the CoS summary achieves a score of 0.80, which is higher than the score of 0.76 achieved using the full source content. This pattern, whereby the summaries created by CoS are more useful than the raw text, is consistent across multiple models and datasets, as demonstrated in Table~\ref{tab:results}. For memorization, the difference is particularly striking with Qwen2.5:7B on TruthfulQA, where CoS achieves 46.06 compared to just 18.55 for zero-shot summaries.

\paragraph{Variability Across Datasets} On \textit{TriviaQA}, all models achieve their highest generalization scores (up to 0.80 with GPT-4o-mini), suggesting its factual nature aligns well with CoS's ability to preserve key information. For \textit{TruthfulQA}, we observe the largest memorization improvements (Qwen2.5:7B increases from 18.55 to 46.06), though generalization scores remain lower, reflecting this dataset's challenge in requiring nuanced understanding of truth claims and its sentence-based answers being harder to evaluate with our Q\&A approach. On \textit{SQuAD 1.1}, while CoS consistently outperforms baselines, overall scores are lower than TriviaQA, likely due to SQuAD's focus on extracting answers from specific contexts rather than testing general knowledge.

\begin{figure*}[htb]
    \centering
    \includegraphics[width=\textwidth]{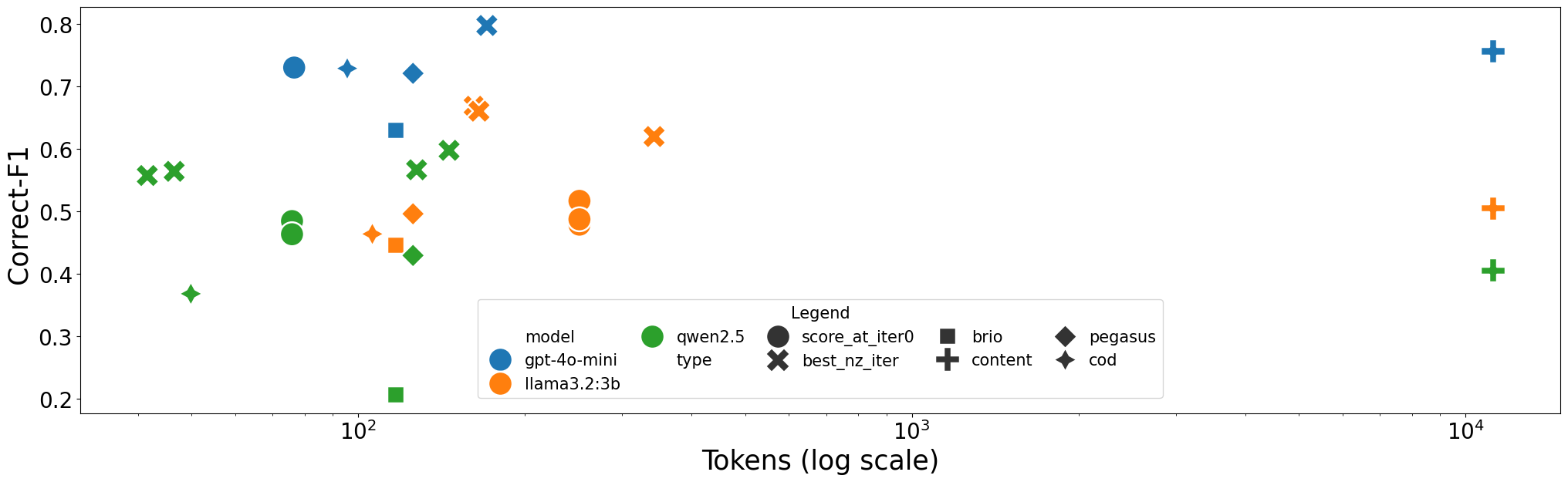}
\caption{Summary length (tokens, log scale) vs. Correct-F1 score on the TriviaQA dataset, comparing performance across model backbones and summarization methods. This visualization demonstrates CoS's ability to create information-dense summaries that optimize the efficiency-performance trade-off across model architectures.}\label{fig:figure2}
\end{figure*}

\paragraph{Synthetic Questions vs. Dataset Questions}
One of the questions for implementing our CoS approach is whether synthetic questions can effectively replace costly human-crafted ones. As shown in Table~\ref{tab:synthetic_vs_train_q}, synthetic questions are indeed viable. Across all three models, synthetic questions yielded comparable or superior results to human-crafted ones. With GPT-4o-mini, synthetic questions improved generalization from 0.73 to 0.80 (+0.07), slightly outperforming human-crafted questions, which improved from 0.74 to 0.79 (+0.05). Similarly, Qwen2.5:7B showed nearly identical improvements with synthetic questions (memorization: +0.29, generalization: +0.12) compared to human-crafted questions (memorization: +0.33, generalization: +0.10). The prompts for question generation and answering can be found in Listings~\ref{lst:qa_generation} and~\ref{lst:answer_prompt}.

\begin{table}[htb]
\centering

\begin{tabular}{lrr}
\toprule
 \textbf{Model} & \textbf{Memorization} & \textbf{Generalization} \\
\midrule
\multicolumn{3}{c}{\emph{Questions from Training Set (human-crafted)}}\\
 GPT-4o-mini & 0.43 $\rightarrow$ 0.51 & 0.74 $\rightarrow$ 0.79 \\
 Llama3.2:3B & 0.21 $\rightarrow$ 0.39 & 0.52 $\rightarrow$ 0.64 \\
 Qwen2.5:7B & 0.23 $\rightarrow$ 0.56 & 0.48 $\rightarrow$ 0.58 \\
\midrule
\multicolumn{3}{c}{\emph{Synthetic Questions}}\\
 GPT-4o-mini & 0.43 $\rightarrow$ 0.48 & 0.73 $\rightarrow$ 0.80 \\
 Llama3.2:3B & 0.21 $\rightarrow$ 0.39 & 0.52 $\rightarrow$ 0.62 \\
 Qwen2.5:7B & 0.23 $\rightarrow$ 0.52 & 0.48 $\rightarrow$ 0.60 \\
\bottomrule
\end{tabular}
\caption{Improvement in F1-score of Chain-of-Summaries against the zero-shot summarization baseline on TriviaQA. Memorization lists the performance on those questions based on which the summary was created (either from the human-crafted set or based on synthetic questions). Generalization lists the performance on the validation set of TriviaQA.
Synthetic questions resemble a viable alternative when no ground-truth questions are available.}\label{tab:synthetic_vs_train_q}
\end{table}

Table~\ref{tab:synthetic-vs-human} contrasts representative synthetic questions produced by our generator with human-authored questions drawn from the TriviaQA evaluation set. The examples are from the same topical neighborhood (Sherlock Holmes) to highlight qualitative differences attributable to question provenance rather than subject matter.

\begin{table}[h!]
\centering
\small
\begin{tabular}{p{0.48\linewidth} p{0.48\linewidth}}
\toprule
\textbf{Synthetic (Generated)} & \textbf{Human (TriviaQA Eval)} \\
\midrule
Who created the character of Sherlock Holmes? \newline \textit{Answer:} Arthur Conan Doyle
&
In the Sherlock Holmes stories who was Moriarty’s second in command? \newline \textit{Answer:} Colonel Moran
\\[0.8em]
In what year did Sherlock Holmes first appear in print? \newline \textit{Answer:} 1887
&
What item of headwear is associated with Sherlock Holmes? \newline \textit{Answer:} Deerstalker hat
\\[0.8em]
Where is Holmes's address located? \newline \textit{Answer:} 221B Baker Street
&
In which London magazine did Sherlock Holmes first appear? \newline \textit{Answer:} The Strand
\\
\bottomrule
\end{tabular}
\caption{Side-by-side examples of synthetic versus human questions in the Sherlock Holmes domain. Synthetic questions tend to be more canonical and factoid-centric, whereas human questions more often reflect culturally mediated associations and narrative-specific details.}
\label{tab:synthetic-vs-human}
\end{table}

\paragraph{Hyperparameter Sensitivity Analysis}
We investigated three key hyperparameters in our CoS approach: number of iterations, questions per iteration, and integration of Chain of Draft, a prompting method where the model uses a reduced number of thinking output tokens. As shown in Table \ref{tab:hyper_param_analysis}, distributing questions over multiple iterations consistently outperforms presenting all questions simultaneously. For Llama3.2:3B, using 10 iterations with 1 question per iteration improved generalization from 0.52 to 0.64, while the single iteration approach with 50 questions reduced performance to 0.49. Similarly, for Qwen2.5:7B, the 10-iteration configuration improved generalization from 0.48 to 0.58, while the single-iteration approach reduced performance to 0.39. The inclusion of Chain of Draft showed model-dependent effects. For Llama3.2:3B, Chain of Draft achieved the highest generalization score of 0.68 with 10 iterations and 1 question per iteration - an improvement of 6\%. However, for Qwen2.5:7B, no Chain of Draft performed better (0.60 versus 0.56). These results highlight the importance of balancing the number of iterations with the number of questions per iteration, and taking into account model-specific characteristics.

\begin{table*}[ht]
\centering
\resizebox{\textwidth}{!}{
\begin{tabular}{lcccccc}
\toprule
\textbf{Method} & \multicolumn{2}{c}{\textbf{T Metrics}} & \multicolumn{2}{c}{\textbf{S Metrics}} & \multicolumn{2}{c}{\textbf{Configuration}} \\
& \textbf{Memorization} & \textbf{Generalization} & \textbf{Memorization} & \textbf{Generalization} & \textbf{Iterations} & \textbf{Q per I} \\
% \midrule
% \multicolumn{7}{c}{\emph{GPT-4o-mini}}\\
% Chain of Summaries & 0.43 $\rightarrow$ 0.51 & 0.74 $\rightarrow$ 0.79 & 0.43 $\rightarrow$ 0.48 & 0.73 $\rightarrow$ 0.80 & 10 & 1 \\
% Chain of Summaries & 0.43 $\rightarrow$ 0.51 & 0.74 $\rightarrow$ 0.79 & 0.43 $\rightarrow$ 0.51 & 0.74 $\rightarrow$ 0.78 & 5 & 10 \\
% Chain of Summaries w/ CoD & 0.43 $\rightarrow$ 0.46 & 0.73 $\rightarrow$ 0.72 & 0.43 $\rightarrow$ 0.48 & 0.74 $\rightarrow$ 0.71 & 1 & 50 \\
\midrule
\multicolumn{7}{c}{\emph{Llama3.2:3B}}\\
Chain of Summaries & 0.21 $\rightarrow$ 0.33 & 0.52 $\rightarrow$ 0.49 & 0.21 $\rightarrow$ 0.32 & 0.52 $\rightarrow$ 0.52 & 1 & 50 \\
Chain of Summaries & 0.21 $\rightarrow$ \textbf{0.40} & 0.52 $\rightarrow$ 0.60 & 0.21 $\rightarrow$ \textbf{0.40} & 0.52 $\rightarrow$ 0.60 & 5 & 10 \\
Chain of Summaries & 0.21 $\rightarrow$ \textit{0.39} & 0.52 $\rightarrow$ \textit{0.64} & 0.21 $\rightarrow$ \textit{0.39 }& 0.52 $\rightarrow$\textit{ 0.62} & 10 & 1 \\
Chain of Summaries w/ Chain of Draft & 0.21 $\rightarrow$ 0.29 & 0.48 $\rightarrow$ 0.45 & 0.21 $\rightarrow$ 0.30 & 0.48 $\rightarrow$ 0.44 & 1 & 50 \\
Chain of Summaries w/ Chain of Draft & 0.21 $\rightarrow$ 0.36 & 0.48 $\rightarrow$ 0.61 & 0.20 $\rightarrow$ 0.34 & 0.48 $\rightarrow$\textit{ 0.62} & 5 & 10 \\
Chain of Summaries w/ Chain of Draft & 0.21 $\rightarrow$ 0.33 & 0.48 $\rightarrow$ \textbf{0.68} & 0.20 $\rightarrow$ 0.33 & 0.48 $\rightarrow$ \textbf{0.67} & 10 & 1 \\
\midrule
\multicolumn{7}{c}{\emph{Qwen2.5:7B}}\\
Chain of Summaries & 0.23 $\rightarrow$ 0.53 & 0.48 $\rightarrow$ 0.39 & 0.23 $\rightarrow$ 0.56 & 0.48 $\rightarrow$ 0.40 & 1 & 50 \\
Chain of Summaries & 0.23 $\rightarrow$ \textbf{0.63} & 0.48 $\rightarrow$ 0.53 & 0.23 $\rightarrow$ \textbf{0.63} & 0.48 $\rightarrow$ 0.53 & 5 & 10 \\
Chain of Summaries & 0.23 $\rightarrow$ 0.56 & 0.48 $\rightarrow$ \textbf{0.58} & 0.23 $\rightarrow$ 0.52 & 0.48 $\rightarrow$ \textbf{0.60} & 10 & 1 \\

Chain of Summaries w/ Chain of Draft & 0.23 $\rightarrow$ 0.49 & 0.47 $\rightarrow$ 0.36 & 0.23 $\rightarrow$ 0.51 & 0.46 $\rightarrow$ 0.37 & 1 & 50 \\
Chain of Summaries w/ Chain of Draft & 0.23 $\rightarrow$ \textit{0.59} & 0.47 $\rightarrow$ 0.47 & 0.23 $\rightarrow$ \textit{0.58} & 0.46 $\rightarrow$ 0.47 & 5 & 10 \\
Chain of Summaries w/ Chain of Draft & 0.23 $\rightarrow$ 0.41 & 0.47 $\rightarrow$ \textit{0.54} & 0.23 $\rightarrow$ 0.41 & 0.46 $\rightarrow$ \textit{0.56} & 10 & 1 \\
\bottomrule
\end{tabular}%
}
\caption{Hyperparameter Sensitivity Analysis of Chain of Summaries Approach on TriviaQA Dataset. This table demonstrates how different configurations of iterations and questions per iteration (Q per I) affect model performance across Llama3.2:3B and Qwen2.5:7B models. Performance is measured using training metrics (T Metrics) and validation metrics (S Metrics), each reporting memorization and generalization capabilities. The results reveal that distributing questions across multiple iterations (10 iterations with 1 question or five iterations with 10 questions) consistently outperforms concentrating all questions in a single iteration (1 iteration with 50 questions). For Llama3.2:3B, Chain of Summaries with Chain of Draft achieves the highest generalization scores (0.68/0.67) when using 10 iterations with 1 question, while Qwen2.5:7B performs best without Chain of Draft. Bold values indicate best overall performance, and italic values show second-best performance for each model and metric combination.}\label{tab:hyper_param_analysis}

\end{table*}

\paragraph{Downstream QA Performance on TriviaQA}

Our Chain of Summaries approach, using a GPT-4o-mini backbone, achieves a competitive 80.0\% F1-score on TriviaQA. This result in Table \ref{tab:triviaqa_results} outperforms context compression methods like RECOMP (59\% F1) and surpasses several sophisticated RAG frameworks, including ATLAS (76.4\%) and REPLUG (77.3\%). CoS is a model-agnostic method that requires no fine-tuning. Since the context summary is generated before the final QA step, CoS adds no compute overhead at the inference stage. While fine-tuned models hold state-of-the-art (e.g., RankRAG at 92.3\%), CoS offers a flexible and efficient black-box alternative that can enhance any general-purpose LLM by solely improving its input context, making it a highly accessible and adaptable solution that could be moderated by humans to ensure quality and safety.

\begin{table}[t]
\centering
\begin{tabular}{lr}
\toprule
\textbf{Model} & \textbf{TriviaQA F1} \\
\midrule
\multicolumn{2}{c}{\emph{Published Results}} \\
RA-DIT (65B) & 39.7 \\
RECOMP (20B) & 59.0 \\
Raven (11B) & 65.7 \\
TriviaQA Atlas (11B) & 74.5 \\
FIT-RAG + Llama2-13B-Chat  & 75.2 \\
RePlug (65B) & 77.3 \\
Llama3-ChatQA-1.5-70B  & 85.6 \\
GenKI & 86.7 \\
Llama3-RankRAG (70B) & 92.3 \\
\midrule
\multicolumn{2}{c}{\emph{Chain of Summaries}} \\
GPT-4o-mini & 80.0 \\
Llama3.2 (3B) & 62.0 \\
Qwen2.5 (7B) & 60.0 \\
\bottomrule
\end{tabular}%
\caption{TriviaQA F1 scores. Our fine-tuning-free CoS method with a GPT-4o-mini backbone achieves 80.0\% F1, outperforming strong retrieval-augmented baselines like ATLAS (74.5\%) and REPLUG (77.3\%).}
\label{tab:triviaqa_results}
\end{table}

\section{Efficiency \& Cost Analysis}

CoS consistently outperforms traditional methods while using substantially fewer tokens (Figure \ref{fig:figure2}). With only 170 tokens, GPT-4o-mini achieves F1=0.80, outperforming zero-shot (0.73), BRIO (0.63), and PEGASUS (0.72). Content-based approaches require over 11,000 tokens for F1=0.76, making them impractical. Llama3.2:3B reaches F1=0.62, exceeding zero-shot (0.52) with similar token efficiency. Qwen2.5 demonstrates the strongest efficiency gains, achieving F1=0.60 using just 42–47 tokens.

However, these efficiency gains come with an important drawback: the CoS approach requires an initial pre-processing step to create the summary structure. Generating a CoS summary incurs a one-time indexing cost that varies with document size. Table~\ref{tab:efficiency} presents a detailed breakdown of token costs across three representative document sizes, split by input and output tokens.

\begin{table}[t]
\centering
\resizebox{\columnwidth}{!}{
\begin{tabular}{llccc}
\toprule
\textbf{Phase} & \textbf{Tokens} & \textbf{5k Doc} & \textbf{10k Doc} & \textbf{50k Doc} \\
\midrule
\multirow{2}{*}{A: Question Generation} & Input & 5,200 & 10,200 & 50,200 \\
 & Output & 300 & 300 & 300 \\
\midrule
\multirow{2}{*}{B: Initial Summary} & Input & 5,200 & 10,200 & 50,200 \\
 & Output & 100 & 170 & 500 \\
\midrule
\multirow{2}{*}{C: Refinement (10 iter.)} & Input & 57,150 & 108,550 & 515,150 \\
 & Output & 3,000 & 3,700 & 7,000 \\
\midrule
\multirow{2}{*}{\textbf{Total Indexing}} & Input & 67,550 & 128,950 & 615,550 \\
 & Output & 3,400 & 4,170 & 7,800 \\
\midrule
\multicolumn{2}{l}{Summary Size} & 100 & 170 & 500 \\
\multicolumn{2}{l}{Savings per Query} & 4,900 & 9,830 & 49,500 \\
\multicolumn{2}{l}{Break-even (queries)} & $\sim$15 & $\sim$14 & $\sim$13 \\
\midrule
\multicolumn{2}{l}{Indexing Cost (USD)} & \$0.012 & \$0.022 & \$0.097 \\
\bottomrule
\end{tabular}
}
\caption{Token costs for CoS indexing across document sizes, split by input and output tokens. USD cost calculated using GPT-4o-mini pricing (\$0.15/M input, \$0.60/M output). Summary sizes scale proportionally with document length.}
\label{tab:efficiency}
\end{table}

Consider a typical 10,000-token document. Phase A (Antithesis) consumes 10,200 input tokens (10,000 document + 200 prompt) and generates 300 output tokens (20 questions $\times$ 15 tokens each). Phase B (Thesis) requires 10,200 input tokens (document + prompt) and produces a 170-token summary. The cost is primarily dominated by Phase C (Synthesis) across 10 refinement iterations. Each iteration consists of two steps: (1) an answering step involving reading the current summary (170 tokens), the question list (300 tokens) and the prompt (200 tokens) to produce answers (200 tokens); and (2) a refining step involving reading the document (10,000 tokens), the current summary (170 tokens) and one unanswered question (15 tokens) to generate an updated summary (170 tokens). This yields approximately 10,855 input tokens and 370 output tokens per iteration, totalling 108,550 input tokens and 3,700 output tokens for 10 iterations.

The total indexing cost of approximately 133,000 tokens (128,950 input + 4,170 output) is offset by substantial inference savings: each subsequent retrieval using the 170-token CoS summary instead of the full 10,000-token document saves 9,830 tokens, a 98.3\% reduction. The break-even point occurs after approximately 14 queries ($133{,}000 \div 9{,}830 \approx 13.5$). As Table~\ref{tab:efficiency} shows, this break-even ratio remains remarkably stable across document sizes (12-15 queries), making CoS ``token positive'' early in the deployment lifecycle for documents accessed repeatedly in retrieval-augmented systems.

\paragraph{Cost Implications}
Using GPT-4o-mini API pricing (\$0.15 per million input tokens, \$0.60 per million output tokens), the per-query cost for a 10k document drops from \$0.0015 (full document) to \$0.000026 (CoS summary), a 98.3\% reduction. For a document retrieved 1,000 times, CoS reduces inference costs from \$1.50 to \$0.026, saving \$1.45 after accounting for the one-time indexing cost of \$0.022. This efficiency gain scales multiplicatively across large document collections: indexing 1,000 documents costs approximately \$22, while saving over \$1,400 across 1,000 queries per document.

\section{Discussion}\label{sec:discussion}

% Summary of results
Our main finding is that CoS creates a textual summary that is more useful to a downstream LLM than the source text by condensing information and removing irrelevant context, suggesting that CoS makes information better accessible for downstream language model use. Moreover, CoS consistently outperforms zero-shot summarization, as well as dedicated summarization methods such as BRIO and PEGASUS.

Our findings are in agreement with recent work showing that increased test-time compute (here, multiple LLM calls) can improve the quality of generated text~\cite{muennighoff2025s1} which would amortize over multiple requests in our envisioned scenario, where website maintainers would use CoS to create information-dense summaries for downstream LLM agents to process. If implemented server-side, which is our primary recommendation, website maintainers can inspect the current summary of their website and/or specific subpages, and ingest manual corrections if needed. If they don't have human-crafted Q\&A pairs, website maintainers can use synthetic Q\&A pairs, which we have shown to perform on par with real Q\&A pairs. In contrast, client-side implementation (e.g., on the site of an LLM agent operating on the web) would induce redundancy as different systems would have to maintain their caches and website maintainers would have little control over how their websites are summarized and whether the cache may have been invalidated. 

% Our attempt at fine-tuning an LLM towards good one-step summarization was less effective than the multi-step summarization. We cannot exclude that this would be different with a larger model or through a more sophisticated reward model, which would change this result. However, as of now, we can conclude from this that the iterative procedure is \emph{necessary} to obtain a high-quality summary. 

The computational demands of large language models carry environmental costs~\cite{strubell-etal-2019-energy}. By reducing per-query token consumption by over 98\%, CoS directly decreases inference energy requirements. For repeatedly-accessed documents, CoS's amortized computational footprint is substantially lower than processing full documents. This reduction translates to lower carbon emissions~\cite{schwartz2020greenai}, while enabling lower latency responses for real-time applications. CoS thus achieves greener AI by reducing inference compute while maintaining downstream task performance. 

In sum, we have shown how Hegel's dialectical method can be leveraged with large language models to obtain a general-purpose summary that anticipates future questions, ready to be used as an LLM-friendly cache for content on the web.

\paragraph{Limitations}
Our selection of datasets mainly consists of question-answering datasets. While we deem this a reasonable proxy for practical applications, it does not involve complex multi-stage synthesis of content on the web from various sources, which would need to be carried out by the client LLM. Additionally, our current evaluation focuses exclusively on single-document summarization; future work should explore multi-document scenarios where iterative reasoning could consolidate knowledge across related documents. More broadly, summarizing multiple possibly interlinked documents on the web for LLM use would be an exciting avenue for future work, testing how such links within the summaries could facilitate navigation of LLMs on the web. Future work may further look at more sophisticated question stratification and fine-tuning larger LLMs.

\section{Conclusion}

% What do we do?
We have introduced Chain of Summaries (CoS) as an iterative refinement method for summarization. We have shown that CoS consistently outperforms zero-shot summarization as well as dedicated summarization approaches.
% Why is it great?
As CoS works with synthetic questions, it is particularly suitable for applications with limited labeled data. Crucially, we showed that these information-dense summaries can even surpass the performance of the original documents, suggesting that CoS makes information better accessible for LLMs.
% Why does it matter?
Our CoS method creates efficient textual summaries that serve as knowledge caches for LLMs, reducing redundant computation in retrieval-augmented systems.

\section{Ethical Considerations}
We don't expect that the proposed method would cause any extra harm compared to LLMs operating on the web on their own. In contrast, if implemented server-side, our approach allows website maintainers to validate what textual representation LLMs process.

% \begin{acks}
% \end{acks}


\appendix

\section{Prompts}

\begin{lstlisting}[caption={Prompt for synthetic Q\&A pairs generation}, label={lst:qa_generation}, breaklines=true, basicstyle=\small\ttfamily]
[
  {
    "content": "Here is the content of the file: {file_content}",
    "role": "system"
  },
  {
    "content": "Generate {number_of_questions} diverse and specific questions in Q: format based on the content. Do not include question numbers. Each question should target important information from the text. Each answer should be concise (word or short phrase) and directly address the question.",
    "role": "user"
  },
  {
    "content": "Ensure answers are brief (a word or short phrase, not a full sentence) and factually accurate based on the text. Format should be: 
        Q: <question>
        A: <answer>",
    "role": "user"
  }
]
\end{lstlisting}

\begin{lstlisting}[caption={Function for answer prompt generation}, label={lst:answer_prompt}, breaklines=true, basicstyle=\small\ttfamily]
[
    {
        "content": "Given a question and content, answer the question as simply as possible! Don't answer in complete sentences; words and phrases are sufficient. Answer the question based on the content provided. If the answer is not present in the content, say \"I don't know.\"",
        "role": "system",
    },
    {
        "content": f"Content: {file_content}", 
        "role": "system"
    },
    {
        "content": f"Question: {question} Answer:",
        "role": "user",
    },
]
\end{lstlisting}
\begin{lstlisting}[caption={Function for summary refinement prompt generation}, label={lst:refine_summary_prompt}, breaklines=true, basicstyle=\small\ttfamily]
[
    {
        "content": "You are an expert text summarizer. Your task is to refine an existing summary to address specific user questions.
        Rules:
        - Include information that directly answers the user's questions
        - Preserve relevant key points from the original summary
        - Return the original summary unchanged if it already contains all necessary information
        - Return the original summary if the questions are not relevant to the text
        - Keep the summary short and concise
        - Don't include questions in the summary.
        - Every time start with: Updated Summary:",
        "role": "system",
    },
    {
        "content": f"Knowledge Base Passage: {passage}", 
        "role": "system"
    },
    {
        "content": f"Existing Summary: {existing_summary}", 
        "role": "user"
    },
    {
        "content": f"Questions to Address: {antithesis_questions}
        Provide an updated summary addressing the questions while maintaining the informational content of the original summary.",
        "role": "user",
    }
]
\end{lstlisting}

\bibliographystyle{ACM-Reference-Format}
\bibliography{sample-base}

\end{document}